\documentclass{article}
\usepackage{ijcai16}

\usepackage{epsfig}
\usepackage{url}
\usepackage{graphicx}
\usepackage{amsmath}
\usepackage{subfigure}
\usepackage{color}
\usepackage{multirow}
\usepackage{footnote}
\usepackage{bm}
\usepackage{tikz}
\usepackage{tablefootnote}
\usepackage[bottom]{footmisc}
\usepackage{afterpage}
\usepackage{float} 

\usetikzlibrary{arrows}
\tikzstyle{int}=[draw, fill=blue!20, minimum size=2em]
\tikzstyle{init} = [pin edge={to-,thin,black}]

\usepackage{times}
\title{Unsupervised Learning on Neural Network Outputs:\\ with Application in Zero-shot Learning}
\author{Yao Lu \\ Aalto University \\ University of Helsinki \\ Helsinki Institute for Information Technology \\ \texttt{yaolubrain@gmail.com} }

\begin{document}

\maketitle

\begin{abstract}
The outputs of a trained neural network contain much richer information than just a one-hot classifier. For example, a neural network might give an image of a dog the probability of one in a million of being a cat but it is still much larger than the probability of being a car. To reveal the hidden structure in them, we apply two unsupervised learning algorithms, PCA and ICA, to the outputs of a deep Convolutional Neural Network trained on the ImageNet of 1000 classes. The PCA/ICA embedding of the object classes reveals their visual similarity and the PCA/ICA components can be interpreted as common visual features shared by similar object classes. For an application, we proposed a new zero-shot learning method, in which the visual features learned by PCA/ICA are employed. Our zero-shot learning method achieves the state-of-the-art results on the ImageNet of over 20000 classes.
\end{abstract}

\section{Introduction}
Recently, Convolutional Neural Network (CNN) \cite{lecun1998gradient} has made significant advances in computer vision tasks such as image classification \cite{ciresan2012multi,krizhevsky2012imagenet,szegedy2015going} , object detection \cite{girshick14CVPR,ren15fasterrcnn} and image segmentation \cite{turaga2010convolutional,long2014fully}. Moreover, CNN also sheds lights on neural coding in visual cortex. In \cite{cadieu2014deep}, it has been shown that a trained CNN rivals the representational performance of inferior temporal cortex on a visual object recognition task. Therefore, investigating the properties of a trained CNN is important for both computer vision applications and discovering the principles of neural coding in the brain.

In \cite{hinton2014distilling}, it is shown that the softmax outputs of a trained neural network contain much richer information than just a one-hot classifier. Such a phenomenon is called \textit{dark knowledge}. For input vector $\mathbf{y}=(y_1,...,y_n)$, which is called logits in \cite{hinton2014distilling}, the softmax function produces output vector $\mathbf{x}=(x_1,...,x_n)$ such that
\begin{align}
x_i = \frac{\exp(y_i/T)}{\sum_j \exp(y_j/T)}
\label{sm}
\end{align}
where $T$ is the temperature parameter.
The softmax function assigns positive probabilities to all classes since $x_i > 0$ for all $i$. Given a data point of a certain class as input, even when the probabilities of the incorrect classes are small, some of them are much larger than the others.
For example, in a 4-class classification task (cow, dog, cat, car), given an image of a dog, while a hard target (class label) is $(0,1,0,0)$, a trained neural network might output a soft target $(10^{-6},0.9,0.1,10^{-9})$. An image of a dog might have small chance to be misclassified as cat but it is much less likely to be misclassified as car. 
In \cite{hinton2014distilling}, a technique called knowledge distillation was introduced to further reveal the information in the softmax outputs. Knowledge distillation raises the temperature $T$ in the softmax function to soften the outputs. For example, it transforms $(10^{-6},0.9,0.1,10^{-9})$ to $(0.015, 0.664, 0.319, 0.001)$ by raising temperature $T$ from 1 to 3. It has been shown that adding the distilled soft targets in the objective function helps in reducing generalization error when training a smaller model of an ensemble of models \cite{hinton2014distilling}.
Therefore, the outputs of a trained neural network are far from one-hot hard targets or random noise and they might contain rich statistical structures.
 
In this paper, to explore the information hidden in the outputs, we apply two unsupervised learning algorithms, Principle Component Analysis (PCA) and Independent Component Analysis (ICA)  to the outputs of a CNN trained on the ImageNet dataset \cite{deng2009imagenet} of 1000 object classes. Both PCA and ICA are special cases of the Factor Analysis model, with different assumptions on the latent variables. Factor Analysis is a statistical model which can be used for revealing hidden factors that underlie a vector of random variables. In the case of CNN for image classification, the neurons or computational units in the output layer of a CNN, as random variables, represent object classes. A latent factor might represent a common visual attribute shared by several object classes. It is therefore desirable to visualize, interpret and make use of the Factor Analysis models learned on the outputs of a trained CNN.

\section{Softmax}
\label{softmax}
\begin{figure}[h!]
\centering
\subfigure[Softmax] {
\includegraphics[scale=0.24]{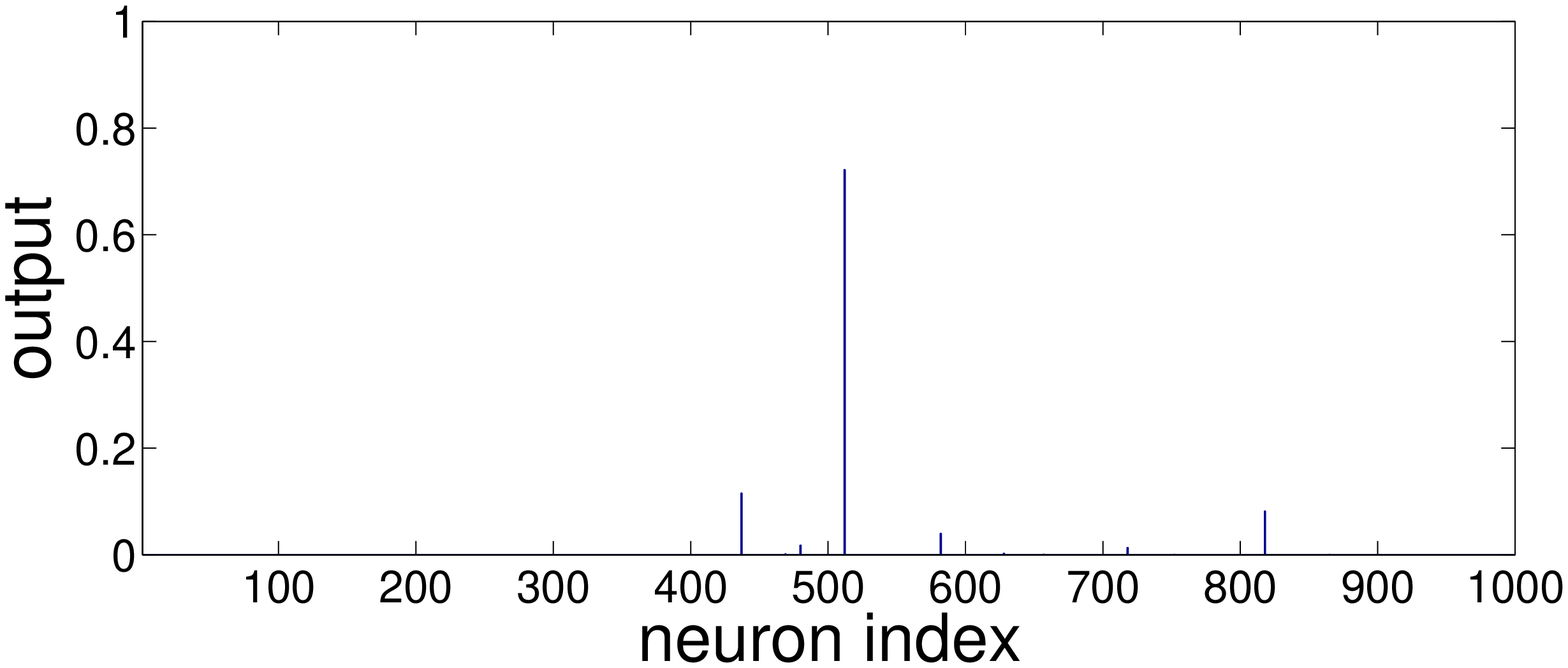}
}
\subfigure[Normalized logits] {
\includegraphics[scale=0.24]{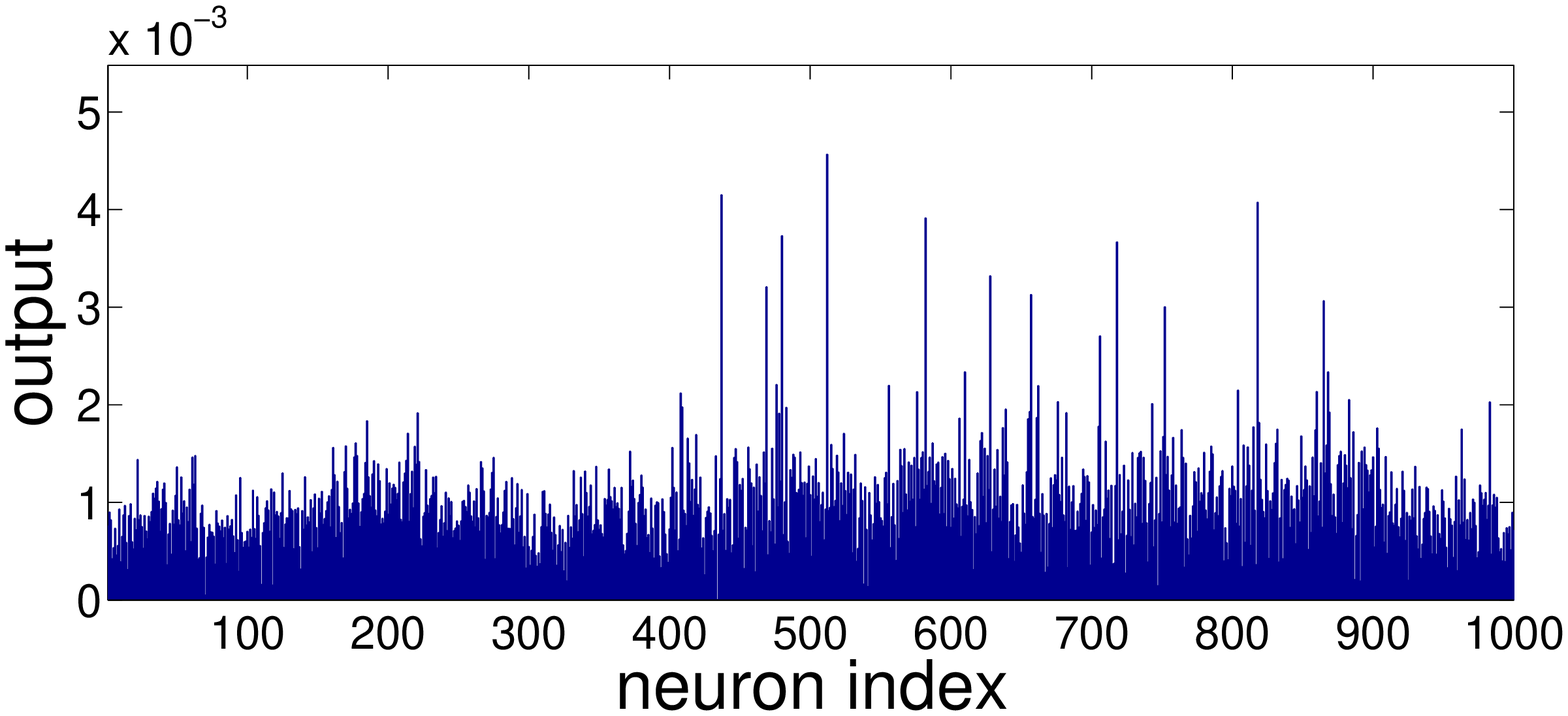}
}
\caption{Outputs}
\label{softmax_fig}
\end{figure}

Because a CNN was trained with one-hot hard targets (class labels), given a training image as input, the softmax function suppresses the outputs of most neurons in the output layer and leaves one or a few peak values. For example, in Figure \ref{softmax_fig} (a), we show the softmax ($T=1$) outputs for a training image. To magnify the  tiny values in the softmax outputs, after a CNN was trained with softmax function ($T=1$), we take the logits $\mathbf{y}$ in e.q. (\ref{sm}) and apply the following normalization function
\begin{align}
x_i = \frac{(y_i - \min_k y_k)}{\sum_j (y_j - \min_k y_k)}
\label{sm_infty}
\end{align}
for all $i$, as the outputs of the CNN, with all the parameters in the CNN unchanged. This function normalizes $\mathbf{y}$ so that $\mathbf{x}$ in eq. (\ref{sm_infty}) is still a probability distribution over classes. We call the $\mathbf{x}$ in eq. (\ref{sm_infty}) normalized logits.
In Figure \ref{softmax_fig} (b), we show the outputs of this function given the same input image as Figure \ref{softmax_fig} (a).

\begin{figure}[h!]
\centering
\subfigure[Softmax] {
\includegraphics[scale=0.24]{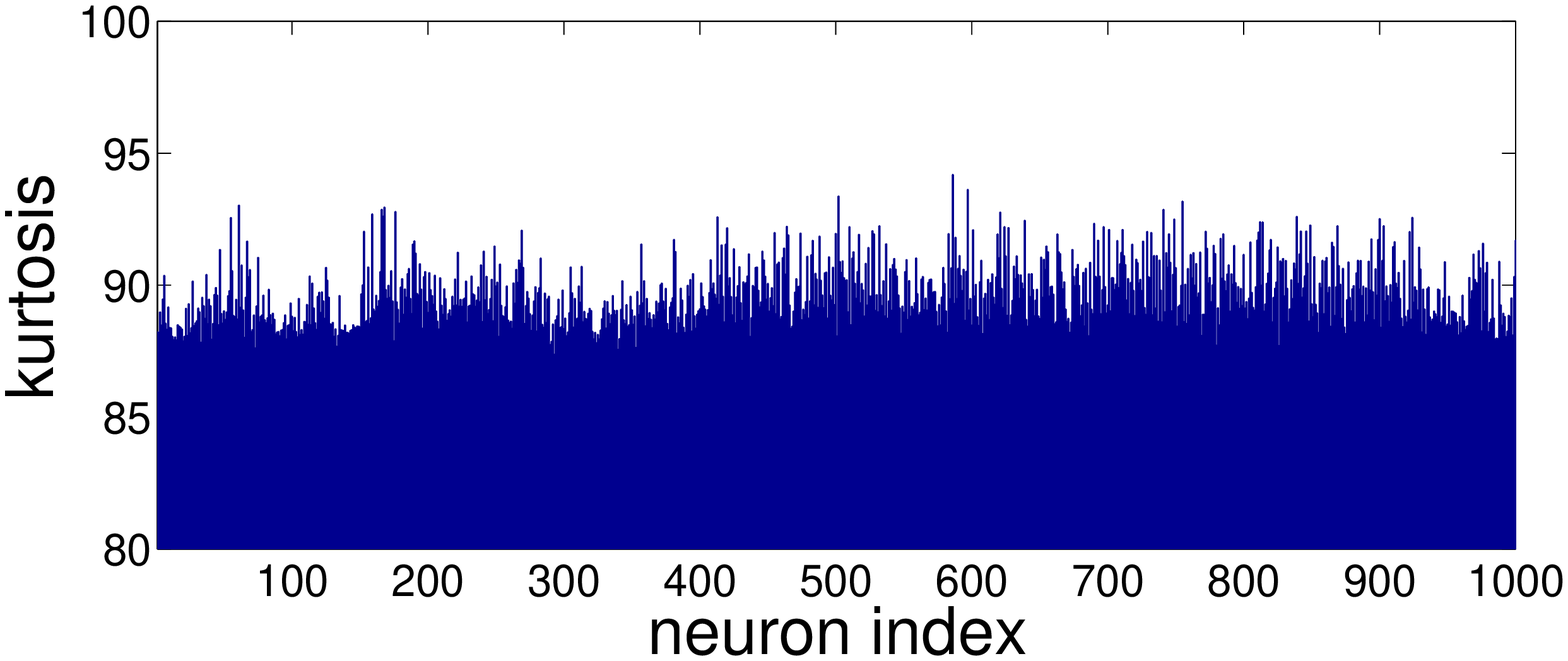}
}
\subfigure[Normalized logits] {
\includegraphics[scale=0.24]{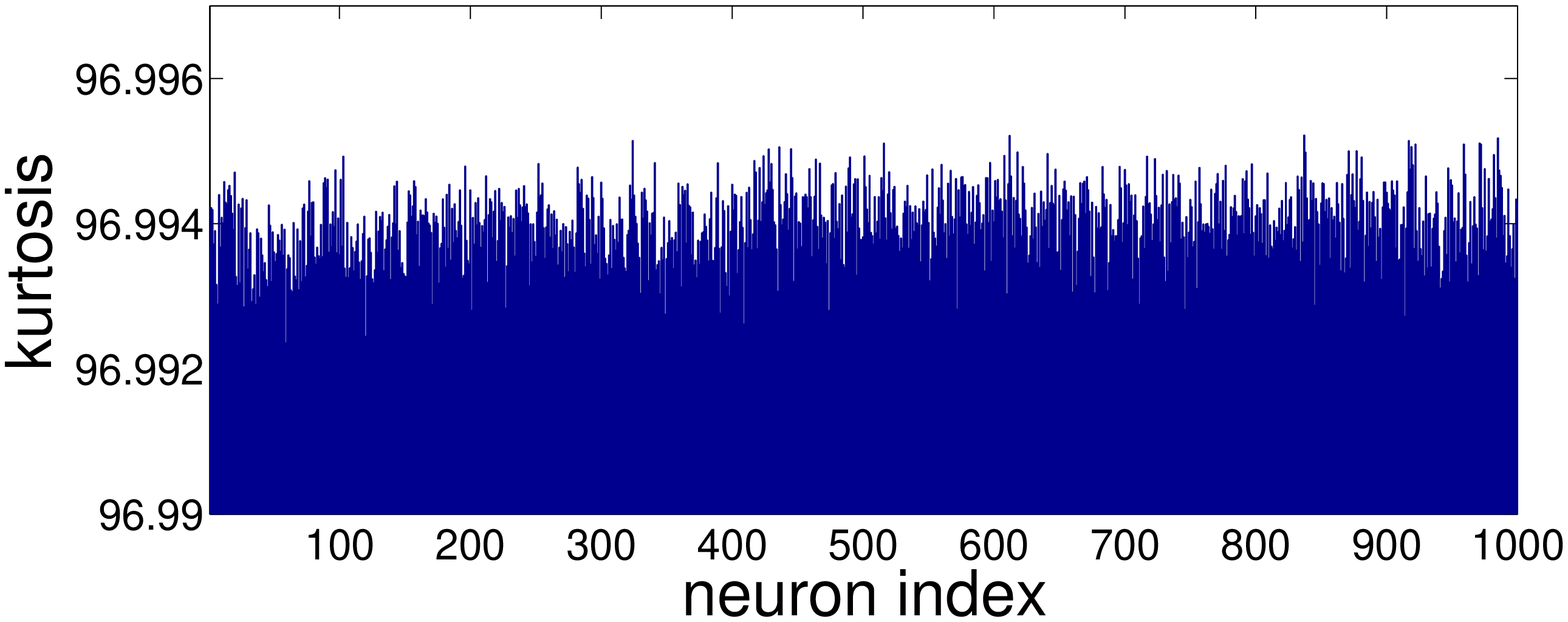}
}
\caption{Kurtosis}
\label{kurtosis}
\end{figure}

In order to apply ICA, the variables must not all be Gaussian. The non-Gaussianity of a random variable $x$ of zero mean can be measured by kurtosis $E(x^4)/E(x^2)^2 -3$, which is zero if $x$ is Gaussian. We computed the kurtosis of the outputs (mean removed) of a CNN  with softmax and normalized logits using all the ImageNet ILSVRC2012 training data. The CNN model and experimental settings are described in Section \ref{results}. The result is, all neurons in the output layer have positive kurtosis, as shown in Figure \ref{kurtosis}. Therefore the neurons as random variables are highly non-Gaussian and it is sensible to apply ICA, which is introduced in the next section.

\section{Factor Analysis}\label{fa}
In Factor Analysis, we assume the observed variables $\mathbf{x} = (x_1,...,x_n)$ are generated by the following model
\begin{align}
\mathbf{x} = \mathbf{As} + \mathbf{n}
\end{align}
where $\mathbf{s}=(s_1,...,s_n)$ are the latent variables, $\mathbf{A}$ is the model parameter matrix and $\mathbf{n}$ are the noise variables. Here, $\mathbf{x}$ and $\mathbf{s}$ are assumed to have zero-mean. $\mathbf{s}$ are also assumed to be uncorrelated and have unit variance, in other words, white.

\subsection{Principle Component Analysis}\label{pca}
Principle Component Analysis (PCA) is a special case of Factor Analysis. In PCA, $\mathbf{s}$ are assumed to be Gaussian and $\mathbf{n}$ are assumed to be zero (noise-free). Let $\mathbf{C}$ denote the covariace matrix of $\mathbf{x}$, $\mathbf{E}=(\mathbf{e}_1,...,\mathbf{e}_n)$ denote the matrix of eigenvectors of $\mathbf{C}$ and $\mathbf{D}=\text{diag}(\lambda_1,...,\lambda_n)$ denote the diagonal matrix of eigenvalues of $\mathbf{C}$. The PCA matrix is $\mathbf{E}^T$, the whitening matrix is $\mathbf{U}=\mathbf{D}^{-1/2}\mathbf{E}^T$ and the  whitened variables are $\mathbf{z} = \mathbf{Ux}$.

\subsection{Independent Component Analysis}\label{ica}
Independent Component Analysis (ICA) \cite{hyvarinen2004independent} is another special case of Factor Analysis. In ICA, $\mathbf{s}$ are assumed to be non-Gaussian and independent and $\mathbf{n}$ are assumed to be zero. ICA seeks a demixing matrix $\mathbf{W}$ such that $\mathbf{Wx}$ can be as independent as possible. To obtain $\mathbf{W}$, we can first decompose it as $\mathbf{W} = \mathbf{VU}$, where $\mathbf{U}$ is the whitening matrix and $\mathbf{V}$ is an orthogonal matrix, which can be learned by maximizing the non-Gaussianity or the likelihood function of $\mathbf{VUx}$. The non-Gaussianity can be measured by kurtosis or negentropy. If dimensionality reduction is required,  we can take the $d$ largest eigenvalues and the corresponding eigenvectors for the whitening matrix $\mathbf{U}$. As a result, the size of $\mathbf{U}$ is $d\times n$ and the size of $\mathbf{V}$ is $d \times d$. Scaling each component does not affect ICA solutions. If $\mathbf{W}$ is an ICA demixing matrix, then $\text{diag}(\alpha_1,...,\alpha_d)\mathbf{W}$ is also an ICA demixing matrix, where $\{\alpha_1,...,\alpha_d\}$ are non-zero scaling constants of the components.

A classic ICA algorithm is FastICA \cite{hyvarinen1999fast}. Despite its fast convergence, FastICA is a batch algorithm which requires all the data to be loaded for computation in each iteration. Thus, it is unsuitable for large scale applications.
To handle large scale datasets, we use a stochastic gradient descent (SGD) based ICA algorithm (described in the Appendix of \cite{hyvarinen1999fast}). For samples $\{\mathbf{z}(1),\mathbf{z}(2),...\}$, one updating step of the SGD-based algorithm of a given sample $\mathbf{z}(t)$ is:
\begin{align}
\mathbf{V} \leftarrow \mathbf{V} + \mu g(\mathbf{V}\mathbf{z}(t))\mathbf{z}(t)^T + \frac{1}{2}(\mathbf{I}-\mathbf{V}\mathbf{V}^T)\mathbf{V}^T
\end{align}
where $\mu$ is the learning rate, $g(\cdot) = -\tanh(\cdot)$ and $\mathbf{I}$ is an identity matrix. In our experiments, $\mathbf{V}$ was initialized as a random orthogonal matrix.

Like FastICA, this SGD-based algorithm requires going through all data once to compute the whitening matrix $\mathbf{U}$. But unlike FastICA, this SGD-based algorithm does not require projection or orthogonalization in each step.

In this algorithm, the assumption on the probability distribution of each $s_i$ is a super-Gaussian distribution
\begin{align}
\log p(s_i) = - \log \cosh(s_i) + \text{constant}
\end{align}
and therefore
\begin{align}
g(s_i) = \frac{\partial}{\partial s_i}\log p(s_i) = -\tanh(s_i).
\end{align}

Since the variables obtained by linear transformations of Gaussian variables are also Gaussian, from Section \ref{softmax}, we can infer at least one neuron in the output layer is non-Gaussian. As an initial attempt, we choose a particular non-Gaussian distribution here. Explorations of different non-Gaussian distributions and therefore different nonlinearities $g(\cdot)$ are left for future research.

\begin{figure}[t!]
\centering
\subfigure[PCA, softmax.] {
\includegraphics[width=\columnwidth]{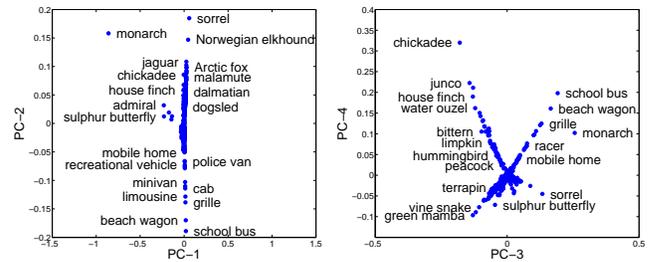}
}
\subfigure[ICA, softmax.] {
\includegraphics[width=\columnwidth]{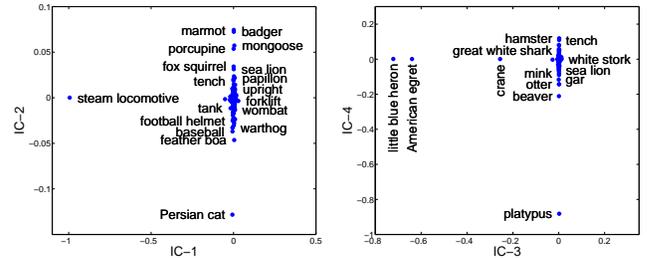}
}
\subfigure[PCA, normalized logits.] {
\includegraphics[width=\columnwidth]{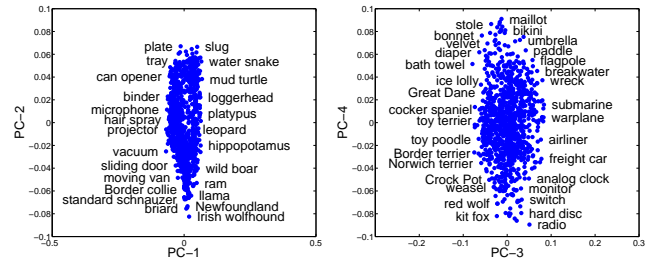}
}
\subfigure[ICA, normalized logits.] {
\includegraphics[width=\columnwidth]{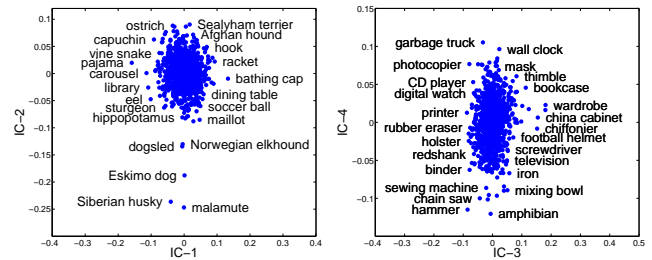}
}
\caption{Label embedding of object class by PCA/ICA components. In each plot, each point is an object class and each axis is a PCA/ICA component (PC/IC). For visual clarity, only selected points are annotated with object class labels.}
\label{ica_plots}
\end{figure}

\section{Results}\label{results}
\begin{table*}[th!]
\centering
\small
\caption{Object classes ranked by single components of PCA/ICA}
\begin{tabular}{|l|c|c|c|c|}
\hline
\rule{0pt}{2ex}   &1 &2 &3 &4  \\
\hline 
\rule{0pt}{2ex}      & mosque & killer whale & Model T & zebra \\
\rule{0pt}{2ex}      & shoji & beaver & strawberry & tiger \\
\rule{0pt}{2ex}PCA   & trimaran & valley & hay & chickadee \\
\rule{0pt}{2ex}      & fire screen & otter & electric locomotive & school bus \\
\rule{0pt}{2ex}      & aircraft carrier & loggerhead & scoreboard & yellow lady's slipper\\
\hline 
\rule{0pt}{2ex}      & mosque & killer whale & Model T & zebra \\
\rule{0pt}{2ex}      & barn & grey whale & car wheel & tiger \\
\rule{0pt}{2ex}ICA   & planetarium & dugong & tractor & triceratops \\
\rule{0pt}{2ex}      & dome & leatherback turtle & disk brake & prairie chicken \\
\rule{0pt}{2ex}      & palace & sea lion & barn & warthog \\
\hline                                                 
\end{tabular}
\label{pca_ica_table}
\end{table*}
\begin{table*}
\centering
\caption{Closest object classes in terms of visual and semantic similarity}
\begin{tabular}{|l|c|c|c|c|}
\hline
\rule{0pt}{2ex}&Egyptian cat &soccer ball &mushroom &red wine  \\
\hline 
\rule{0pt}{2ex}             &tabby cat & rugby ball & bolete & wine bottle\\
\rule{0pt}{2ex}             &tiger cat & croquet ball & agaric  & beer glass \\
\rule{0pt}{2ex}Visual       &tiger & racket & stinkhorn & goblet \\
\rule{0pt}{2ex}             &lynx & tennis ball & earthstar & measuring cup \\
\rule{0pt}{2ex}             &Siamese cat & football helmet & hen-of-the-woods & wine bottle \\
\hline 
\rule{0pt}{2ex}             &Persian cat &croquet ball & cucumber &eggnog\\
\rule{0pt}{2ex}             &tiger cat &golf ball & artichoke &cup\\
\rule{0pt}{2ex}Semantic     &Siamese cat &baseball & cardoon &espresso\\
\rule{0pt}{2ex}             &tabby cat &ping-pong ball & broccoli &menu\\
\rule{0pt}{2ex}             &cougar &punching bag & cauliflower &meat loaf\\   
\hline                                                 
\end{tabular}
\label{vs_table}
\end{table*}

\subsection{Experimental Settings}
\label{experimental_settings}
For the trained CNN model, we used GoogLeNet \cite{szegedy2015going} and AlexNet \cite{krizhevsky2012imagenet}. The results of using two different CNN models are similar. Therefore, due to the space limitation, we only report the results of using GoogLeNet.
We used all the images in the ImageNet ILSVRC2012 training set to compute the ICA matrix using our SGD-based algorithm with mini-batch size 500. The learning rate was set to 0.005 and was halved every 10 epochs. The computation of CNN outputs was done with Caffe \cite{jia2014caffe}. The ICA algorithm was ran with Theano~\cite{bergstra+al:2010-scipy}.

\subsection{Visualization of PCA/ICA components}\label{pca_ica}

To understand what is learned by PCA and ICA, we visualize the PCA and the ICA matrices. In the PCA matrix $\mathbf{E}^T$ or ICA matrix $\mathbf{W}$,  each row corresponds to a PCA/ICA component and each column corresponds to an object class. The number of rows depends on the dimensionality reduction. The number of columns of $\mathbf{E}^T$ or $\mathbf{W}$ is 1000, corresponding to 1000 classes. After the ICA matrix was learned, each ICA component (a row of $\mathbf{W}$) was scaled to have unit $l_2$ norm. The scaling of each ICA component does not affect the ICA solution, as discussed in Section \ref{ica}.

In Figure \ref{ica_plots}, we show the embedding of class labels by PCA and ICA. The horizontal and the vertical axes are two distinct rows of $\mathbf{E}^T$ or $\mathbf{W}$. Each point in the plot corresponds to an object class. And there are 1000 points in each plot. Dimensionality is reduced from 1000 to 200 in ICA. 
In Figure \ref{ica_plots} (a) and (b), we plot two pairs of the PCA/ICA components, learned with softmax outputs. In the PCA embedding, visually similar class labels are along some lines, but not the axes, while in the ICA embedding, they are along the axes. However, most points are clustered in the origin. In Figure \ref{ica_plots} (c) and (d), we plot two pairs of the PCA/ICA components, learned with normalized logits outputs. We can see the class labels are more scattered in the plots.

\begin{figure}[h!]
\centering
\subfigure[PCA] {
\includegraphics[width=\columnwidth]{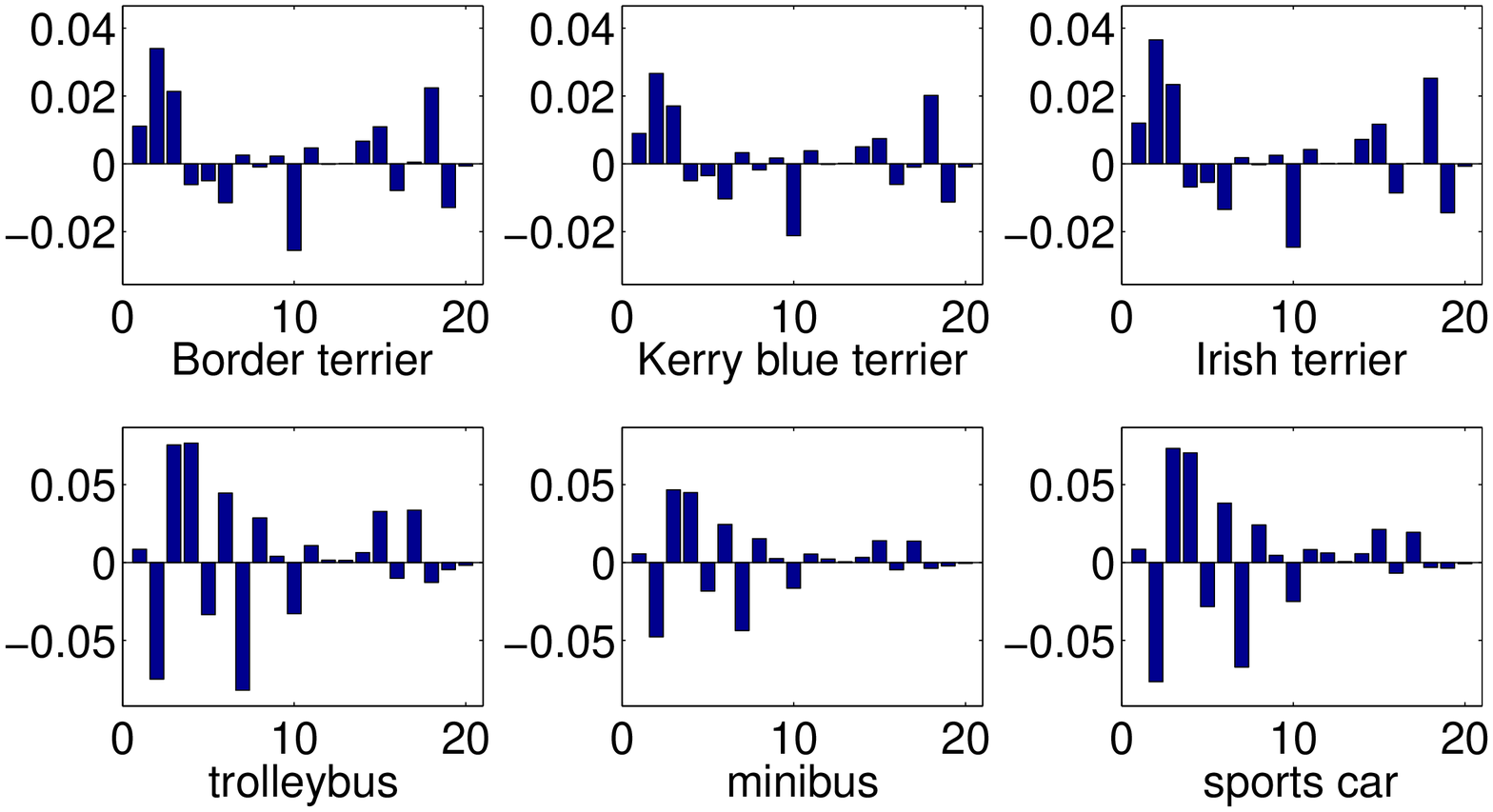}
}
\subfigure[ICA] {
\includegraphics[width=\columnwidth]{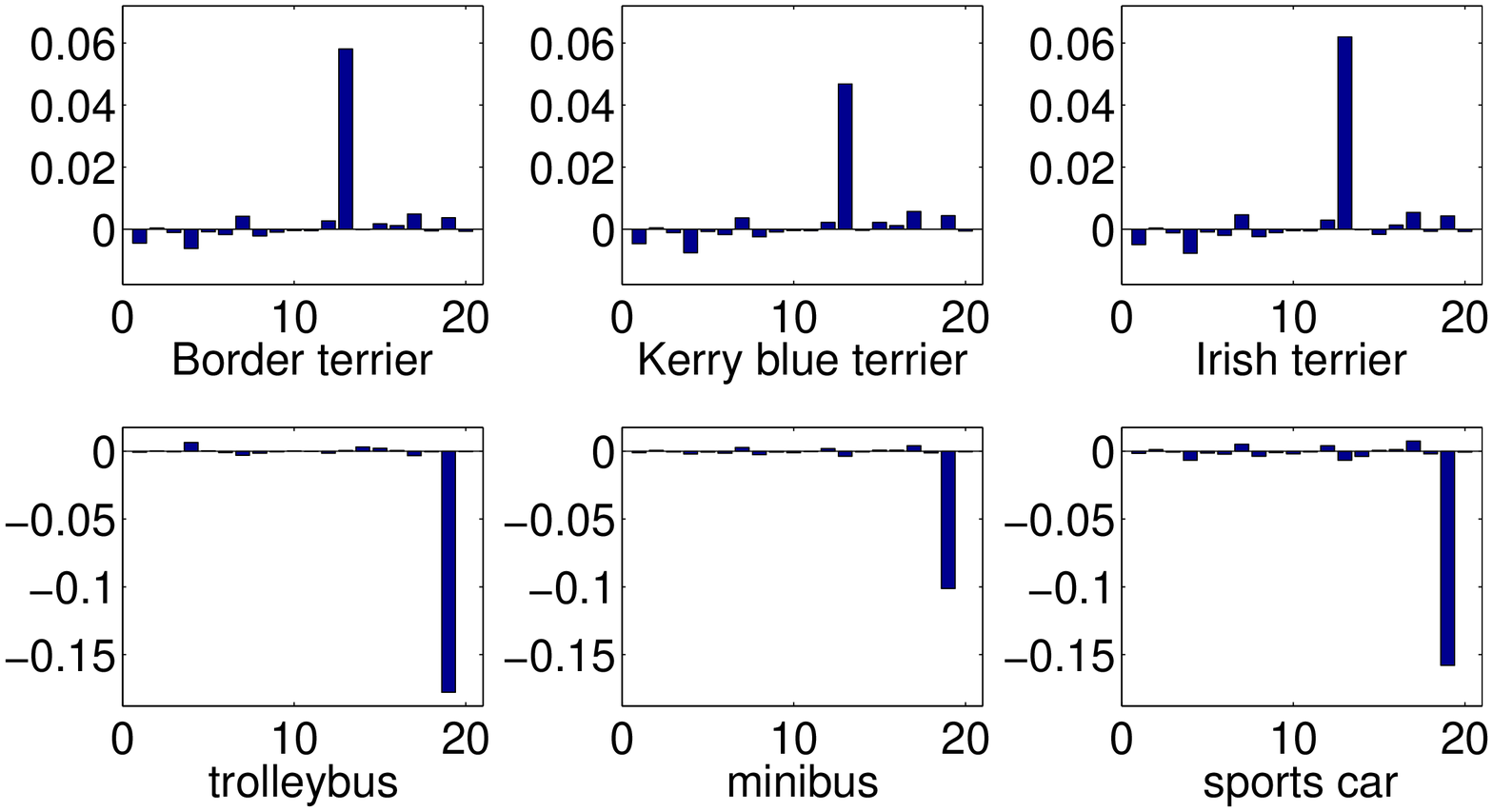}
}
\caption{Bar plots of PCA/ICA componoments of object classes. Dimensionality was reduced to 20 for better visualization.}
\label{bar_plots}
\end{figure}

In Figure \ref{bar_plots}, we show the PCA/ICA componenents of two sets of similar object classes: (1) \textit{Border terrier}, \textit{Lerry blue terrier}, and \textit{Irish terrier}. (2) \textit{trolleybus}, \textit{minibus}, and \textit{sports car}. Both PCA and ICA were learned on the softmax outputs and the dimensionality were reduced to 20 for better visualization. In Figure \ref{bar_plots} (a), we see the PCA components of the object classes are distributed. While in Figure \ref{bar_plots} (b), we see clearly some single components of ICA dominating. There are components representing "dog-ness" and "car-ness". Therefore, the ICA components are more interpretable. 

In Table \ref{pca_ica_table}, we show the top-5 object classes according to the value of PCA/ICA components. For the ease of comparison, we selected each PCA/ICA component which has the largest value for class \textit{mosque}, \textit{killer whale}, \textit{Model T} or \textit{zebra} among all components. We can see that the class labels ranked by ICA components are more visually similar and consistent than the ones by PCA components.

The PCA/ICA components can be interpreted as common features shared by visually similar object classes. From Figure \ref{ica_plots} and Table \ref{pca_ica_table}, we can see the label embeddings of object classes by PCA/ICA components are meaningful since visually similar classes are close in the embeddings. Unlike \cite{akata2013label}, these label embeddings can be unsupervisedly learned with a CNN trained with only one-hot class labels and without any hand annotated attribute label of the object classes, such as \textit{has tail} or \textit{lives in the sea}.

\subsection{Visual vs. Semantic Similarity}\label{vis_sem}

The visual-semantic similarity relationship was previously explored in \cite{deselaers2011visual}, which shows some consistency between two similarities.  Here we further explore it from another perspective. We define the visual and the semantic similarity in the following way. 
The visual similarity between two object classes is defined as cosine similarity of their PCA or ICA components (200-dim and learned with softmax), both of which give the same results. The semantic similarity is defined based on the shortest path length \footnote{Computed with the path\_similarity() function in the NLTK tool \url{http://www.nltk.org/howto/wordnet.html}.} between two classes on the WordNet graph~\cite{fellbaum1998wordnet}. 

In Table \ref{vs_table}, we compare five closest classes of \textit{Egyptian cat}, \textit{soccer ball}, \textit{mushroom} and \textit{red wine} in terms of visual and semantic similarities. For \textit{Egyptian cat}, both visual and semantic similarities give similar results.
For \textit{soccer ball}, \textit{football helmet} is close in terms of visual similarity but distant in terms of semantic similarity. 
For \textit{mushroom} and \textit{red wine}, two similarities give very different closest object classes. The gap between two similarities is intriguing and therefore worth further exploration. In neuroscience literature, it is claimed that visual cortex representation favors visual rather than semantic similarity \cite{baldassi2013shape}.

\newpage

\section{Application: Zero-shot Learning}
To demonstrate the effectiveness of the visual features of object classes learned by PCA and ICA, we apply them to zero-shot learning.
Zero-shot learning \cite{larochelle2008zero,lampert2009learning,palatucci2009zero,rohrbach2011evaluating,socher2013zero} is a classification task in which some classes have no training data at all. We call the classes which have training data seen classes and those which have no training data unseen classes. One can use external knowledge of the classes, such as attributes, to build the relationship between the seen and the unseen classes. Then one can extrapolate the unseen classes by the seen classes. 

Note that the focus of this paper is not zero-shot learning, but the visual features learned by PCA and ICA on the CNN outputs. Our purpose here is to give an example of how PCA and ICA features can be used for computer vision applications. 
Therefore, we do not intend to provide a comprehensive  comparison or review of different zero-shot learning methods. 

\subsection{Previous Work}
Previous state-of-the-art large scale zero-shot learning methods are DeViSE \cite{frome2013devise} and conSE \cite{norouzi2013zero}. Both of them use the ImageNet of 1000 classes for training and the ImageNet of over 20000 classes for testing.
 
In DeViSE, a CNN is first pre-trained on the ImageNet of 1000 classes. Then, 500-dimensional semantic features of both seen and unseen classes are obtained by running word2vec \cite{mikolov2013efficient} on Wikipedia. After that, the last (softmax) layer of the CNN is removed and all the other parts of the CNN are fun-tuned to predict the semantic features of the seen classes for each training image. In testing, when a new image arrives, the prediction is done by computing the cosine similarity of the CNN output vector and the semantic features of classes. In \cite{frome2013devise}, it has also been shown that DeViSE could give more semantically reasonable errors for the seen classes.

In conSE, a CNN is first trained on the ImageNet of 1000 classes and 500-dimensional semantic features of the classes are obtained by running word2vec on Wikipedia, as in DeViSE. However, conSE does not require fun-tuning the CNN to predict the semantic features. The output vector in conSE is a convex combination of the semantic features, by the top activated neurons in the softmax layer. Its testing procedure is the same as DeViSE. In \cite{norouzi2013zero}, it has been shown that conSE gives better performance than DeViSE in the large scale zero-shot learning experiments.

Our method differs from DeViSE and conSE by using unsupervised learning algorithms to learn: (1) visual features of classes. (2) a semantic features of classes from the WordNet graph, instead of Wikipedia. (3) a bridge between the visual and the semantic features.

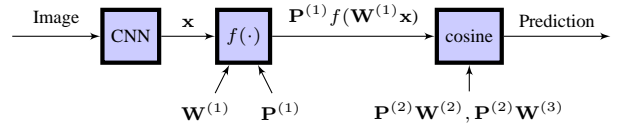
\begin{figure}[t!]
Learning
\begin{center}
\begin{tikzpicture}[node distance=2.5cm,auto,>=latex']
    \node [] (a) at (-2.5,0) {};
    \node [] (b) at (0,-1) {};        
	\node [align=center,int,ultra thick] (c) at (0,0) {\scriptsize{PCA/ICA}};
	\node [] (d) at (2,1) {\scriptsize{$\mathbf{M}$}};	
	\node [align=center,int,ultra thick] (e) at (2,0) {\scriptsize{$f(\cdot)$}};	
	\node [align=center,int,ultra thick] (f) at (2,-1) {\scriptsize{MDS}};		
    \node [align=center,int,ultra thick] (g) at (4.2,-0.5) {\scriptsize{CCA}};		
    \node [] (h) at (5.7,0) {};		
    \node [] (i) at (5.7,-1) {};		
    \node [] (j) at (4.2,-1.5) {};		    
    \path[->] (a) edge node {\scriptsize{CNN outputs}} (c);	
    \path[->] (b) edge node {\scriptsize{WordNet}} (f);	
    \path[->,style={sloped,anchor=south}] (c) edge node {\tiny{$\mathbf{W}^{(1)}$}} (e);	
    \path[->,style={sloped,anchor=south}] (d) edge node {} (e);	
    \path[->,style={sloped,anchor=south}] (e) edge node {\tiny{$f(\mathbf{W}^{(1)}M)$}} (g);	            
    \path[->,style={sloped,anchor=south}] (f) edge node {\tiny{$\mathbf{W}^{(2)}$}} (g);	        
    \path[->,style={sloped,anchor=south}] (f) edge node {\tiny{$\mathbf{W}^{(3)}$}} (j);	            
    \path[->,style={sloped,anchor=south}] (g) edge node {\tiny{$\mathbf{P}^{(1)}$}} (h);	        
    \path[->,style={sloped,anchor=south}] (g) edge node {\tiny{$\mathbf{P}^{(2)}$}} (i);	        
\end{tikzpicture}
\end{center}
Testing
\begin{center}
\begin{tikzpicture}[node distance=2.5cm,auto,>=latex']
    \node [] (a) at (-3.2,0) {};
    \node [] (b) at (0,-1) {};        
	\node [align=center,int,ultra thick] (c) at (-1.5,0) {\scriptsize{CNN}};
	\node [] (d) at (3,-1) {\scriptsize{$\mathbf{P}^{(2)}\mathbf{W}^{(2)}, \mathbf{P}^{(2)}\mathbf{W}^{(3)}$}};	
    \node [align=center,int,ultra thick] (e) at (3,0) {\scriptsize{cosine}};		
    \node [] (f) at (5,0) {};		    
    \node [align=center,int,ultra thick] (g) at (0,0) {\scriptsize{$f(\cdot)$}};		    
    \node [] (h) at (-0.5,-1) {\scriptsize{$\mathbf{W}^{(1)}$}};		        
    \node [] (i) at (0.5,-1) {\scriptsize{$\mathbf{P}^{(1)}$}};		            
    \path[->,style={sloped,anchor=south}] (a) edge node {\scriptsize{Image}} (c);	
    \path[->,style={sloped,anchor=south}] (c) edge node {\scriptsize{$\mathbf{x}$}} (g);	    
    \path[->,style={sloped,anchor=south}] (g) edge node {\scriptsize{$\mathbf{P}^{(1)}f(\mathbf{W}^{(1)}\mathbf{x}$)}} (e);	
    \path[->,style={sloped,anchor=south}] (d) edge node {} (e);	    
    \path[->,style={sloped,anchor=south}] (e) edge node {\scriptsize{Prediction}} (f);	    
    \path[->,style={sloped,anchor=south}] (h) edge node {} (g);	        
    \path[->,style={sloped,anchor=south}] (i) edge node {} (g);	            
\end{tikzpicture}
\end{center}
\caption{Our zero-shot learning method. $\mathbf{W}^{(1)}$ are the visual features of the seen classes. $\mathbf{W}^{(2)}$ are the semantic features of the seen classes. $\mathbf{W}^{(3)}$ are the semantic features of the unseen classes. $\mathbf{P}^{(1)}$ is the projection matrix from visual space to the common space. $\mathbf{P}^{(2)}$ is the projection matrix from semantic space to the common space. $f(\cdot)$ is the $l_1$ normalization. $\mathbf{M}$ are the mean vectors of the seen classes. $\mathbf{x}$ is the CNN output vector.}
\end{figure}

\subsection{Our Method}
Our method works as follows. In the learning phase, first assume we have obtained the visual feature vectors $\mathbf{W}^{(1)}=(\mathbf{w}^{(1)}_1,...,\mathbf{w}^{(1)}_n)$ of $n$ seen classes. Let $\mathbf{M}=(\mathbf{m}_1,...,\mathbf{m}_n)$ denotes the matrix of the mean outputs of a CNN of the seen classes. And $\mathbf{F} = f(\mathbf{W}^{(1)}\mathbf{M}) = (\mathbf{f}_1,...,\mathbf{f}_n)$ are the transformed mean outputs of the seen classes, where $f(\cdot)$ is a nonlinear function. Next, assume we have obtained the semantic feature vectors $\mathbf{W}^{(2)}=(\mathbf{w}^{(2)}_1,...,\mathbf{w}^{(2)}_n)$ of $n$ seen classes and $\mathbf{W}^{(3)}=(\mathbf{w}^{(3)}_1,...,\mathbf{w}^{(3)}_m)$ of $m$ unseen classes. 
Due to the visual-semantic similarity gap shown in Section \ref{vis_sem}, we learn a bridge between the visual and the semantic representations of object classes via Canonical Correlation Analysis (CCA) \cite{hotelling1936relations,hardoon2004canonical}, which seeks two projection matrices $\mathbf{P}^{(1)}$ and $\mathbf{P}^{(2)}$ such that
\begin{align}
\min_{\mathbf{P}^{(1)},\mathbf{P}^{(2)}}\| \mathbf{P}^{(1)T}\mathbf{F} - \mathbf{P}^{(2)T}\mathbf{W}^{(2)} \|_F \\
\text{s.t.} \quad \mathbf{P}^{(k)T} \mathbf{C}_{kk}\mathbf{P}^{(k)} = \mathbf{I},\quad
\mathbf{p}_{i}^{(k)T}\mathbf{C}_{kl}\mathbf{p}_{j}^{(l)} = 0, \\
k,l = 1,2, \quad k \neq l, \quad i,j = 1,...,d,
\end{align}
where $\mathbf{p}_{i}^{(k)}$ is the $i$-th column of $\mathbf{P}^{(k)}$ and $\mathbf{C}_{kl}$ is a covariance or cross-covariance matrix of $\{\mathbf{f}_1,...,\mathbf{f}_n\}$ and/or $\{\mathbf{w}^{(2)}_1,...,\mathbf{w}^{(2)}_n\}$.  

In the testing phase, when a new image arrives, we first compute its CNN output $\mathbf{x}$. Then for $\mathbf{P}^{(1)T} ( f(\mathbf{W}^{(1)}\mathbf{x}) - \frac{1}{n}\sum_i\mathbf{f}_i )$, we compute its $k$ closest columns of $\mathbf{P}^{(2)T}\mathbf{W}^{(2)}$ (seen) and/or $\mathbf{P}^{(2)T}\mathbf{W}^{(3)}$ (unseen).
The corresponding classes of these $k$ columns are the top-$k$ predictions. The closeness is measured by cosine similarity. 

For $\mathbf{W}^{(1)}$, we compare random, PCA, and ICA matrices of different dimensionality in our experiments. The random matrices are semi-orthogonal, that is, $\mathbf{W}^{(1)}\mathbf{W}^{(1)T}=\mathbf{I}$ but $\mathbf{W}^{(1)}\mathbf{W}^{(1)T}\neq \mathbf{I}$. For $\mathbf{W}^{(2)}$ and $\mathbf{W}^{(3)}$, we use the feature vectors by running classic Multi-dimensional Scaling (MDS) on a distance matrix of both seen and unseen classes. The distance between two classes is measured by one minus the similarity in Section \ref{vis_sem}. Each column of $\mathbf{W}^{(2)}$ and $\mathbf{W}^{(3)}$ is subtracted by $\frac{1}{n}\sum_i\mathbf{w}^{(2)}_i$. $\mathbf{M}$ is approximated by $\mathbf{I}$ and $f(\cdot)$ is the scaling normalization of a vector or each column of a matrix to unit $l_1$ norm. We experimented with softmax with different $T$ and normalized logits as the outputs. The best performance (as in Table \ref{zero_shot_unseen}, \ref{zero_shot_images}, \ref{zero_shot_seen}) was obtained with the softmax ($T=1$) output for $\mathbf{x}$ but $\mathbf{E}^T$ and $\mathbf{W}$ were learnt with normalized logits.

In our method, instead of using word2vec on Wikipedia as in DeViSE and conSE, we use classic MDS of the WordNet distance matrix to obtain the semantic features of classes, for simplicity. 
Word embedding on Wikipedia typically consumes a large amount of RAM and takes hours for computation. While classic MDS on the WordNet distance matrix of size 21842$\times$21842\footnote{21841 classes in ImageNet 2011fall plus class \textit{teddy, teddy bear}. Class \textit{teddy, teddy bear} (WordNet ID: n04399382) is in ImageNet ILSVRC2012 but not in ImageNet 2011fall.}  is much cheaper to compute. The computation of a 21632-dimensional MDS feature vector for each class was done in MATLAB with 8 Intel Xeon 2.5GHz cores within 12 minutes. A comprehensive comparisons of different semantic features of classes for zero-shot learning can be found in \cite{akata2015evaluation}.

\subsection{Experiments}
Following the zero-shot learning experimental settings of DeViSE and conSE, we used a CNN trained on ImageNet ILSVRC2012 (1000 seen classes), and test our method to classify images in ImageNet 2011fall (20842 unseen classes~\footnote{Since class \textit{teddy, teddy bear} is missing in ImageNet 2011fall, the correct number of classes is 21841$-$(1000$-$1) = 20842 rather than 20841.}, 21841 both seen and unseen classes). We use top-$k$ accuracy (also called flat hit@$k$ in \cite{frome2013devise,norouzi2013zero}) measure, the percentage of test images in which a method's top-$k$ predictions return the true label. 

For the trained CNN model, we experimented with GoogLeNet and AlexNet.
Although GoogLeNet outperformans AlexNet on the seen classes, our method with the two different CNN models performans essentially the same on the zero-shot learning tasks. Due to the space limitation, we only report the results of using GoogLeNet.

The sizes of the matrices in our methods: $\mathbf{W}^{(1)}$ is $k\times$1000, $\mathbf{W}^{(2)}$ is 21632$\times$1000, $\mathbf{W}^{(3)}$ is 21632$\times$20842, $\mathbf{P}^{(1)}$ is $k\times k$, $\mathbf{P}^{(2)}$ is $k\times$21632, $\mathbf{M}$ is 1000$\times$1000 and $\mathbf{x}$ is 
1000$\times$1. We used $k=100,500,900$ in our experiments. Although $\mathbf{W}^{(2)}$ and $\mathbf{W}^{(3)}$ are large matrices, we only need to compute once and store $\mathbf{P}^{(2)}\mathbf{W}^{(2)}$ and $\mathbf{P}^{(2)}\mathbf{W}^{(3)}$ of size $k\times$1000 and $k\times$21632, respectively.

\begin{table*}[t!]
\centering
\scriptsize
\begin{minipage}{\textwidth}

\begin{center}

\caption{Predictions of test images of unseen classes (correct class labels are in blue)}
\label{zero_shot_images}
\begin{tabular}{|c|c|c|c|}
\hline
\rule{0pt}{3ex} {\small Test Images} & {\small DeViSE~\cite{frome2013devise}} & {\small ConSE~\cite{norouzi2013zero}} & {\small Our Method} \\
\hline
\rule{0pt}{2.5ex}\multirow{5}{*}{\includegraphics[width=0.66in,height=0.66in]{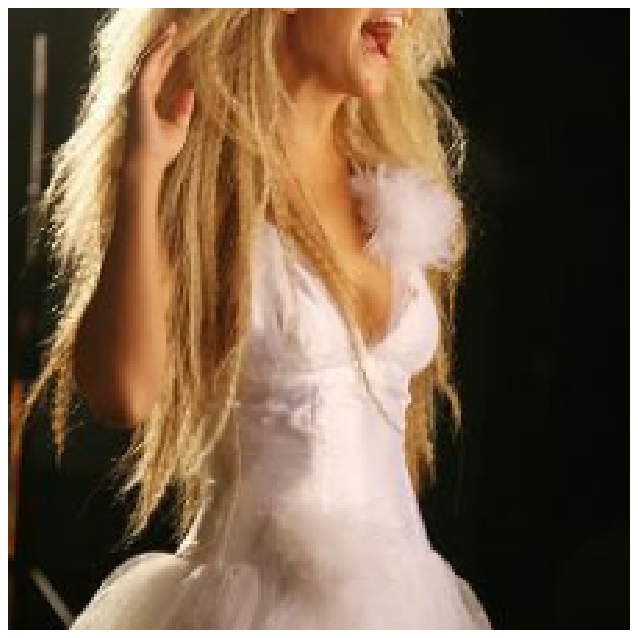}} 
\rule{0pt}{2.5ex}& water spaniel & business suit & periwig, peruke\\
\rule{0pt}{2.5ex}& tea gown & \textbf{\color{blue}dress, frock} & horsehair wig \\
\rule{0pt}{2.5ex}& bridal gown, wedding gown & hairpiece, false hair, postiche & hound, hound dog \\
\rule{0pt}{2.5ex}& spaniel & swimsuit, swimwear, bathing suit & bonnet macaque \\
\rule{0pt}{2.5ex}& tights, leotards & kit, outfit & toupee, toupe \\
\hline
\rule{0pt}{2.5ex}\multirow{5}{*}{\includegraphics[width=0.66in,height=0.66in]{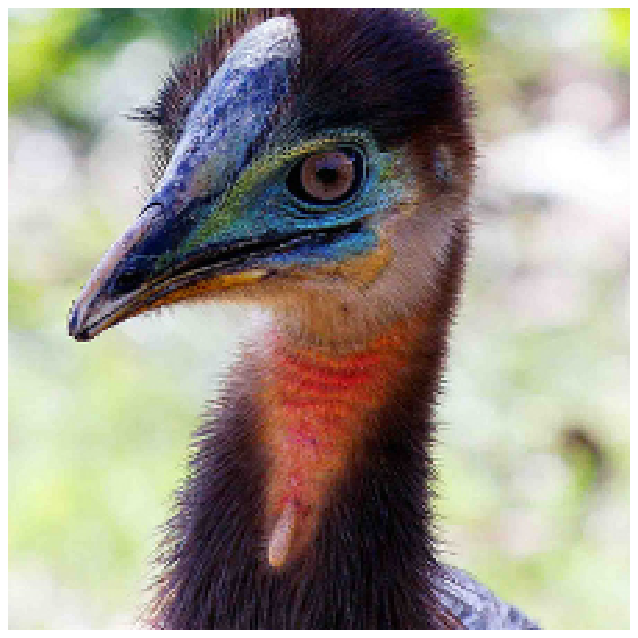}} 
\rule{0pt}{2.5ex}&heron & \textbf{\color{blue}ratite, ratite bird, flightless bird} & \textbf{\color{blue}ratite, ratite bird, flightless bird} \\
\rule{0pt}{2.5ex}&owl, bird of Minerva, bird of night & peafowl, bird of Juno & kiwi, apteryx \\
\rule{0pt}{2.5ex}&hawk & common spoonbill & moa \\
\rule{0pt}{2.5ex}&bird of prey, raptor, raptorial bird & New World vulture, cathartid & elephant bird, aepyornis \\
\rule{0pt}{2.5ex}&finch & Greek partridge, rock partridge & emu, Dromaius novaehollandiae\\
\hline
\rule{0pt}{2.5ex}\multirow{5}{*}{\includegraphics[width=0.66in,height=0.66in]{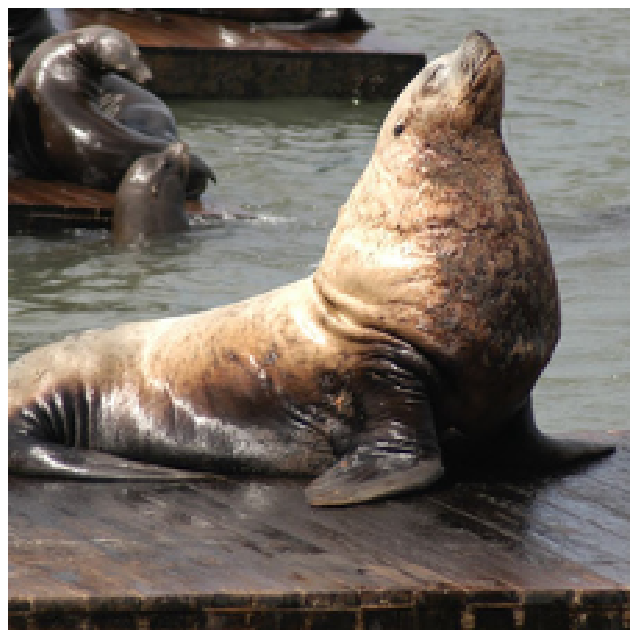}} 
\rule{0pt}{2.5ex}&elephant & California sea lion & fur seal \footnote{WordNet ID: n02077152. There are two classes named \textit{fur seal} with different WordNet IDs.
} \\ 
\rule{0pt}{2.5ex}&turtle & \textbf{\color{blue}Steller sea lion} & eared seal\\ 
\rule{0pt}{2.5ex}&turtleneck, turtle, polo-neck & Australian sea lion & fur seal \footnote{WordNet ID: n02077658.} \\
\rule{0pt}{2.5ex}&flip-flop, thong & South American sea lion & guadalupe fur seal\\
\rule{0pt}{2.5ex}&handcart, pushcart, cart, go-cart & eared seal & Alaska fur seal\\
\hline
\rule{0pt}{2.5ex}\multirow{5}{*}{\includegraphics[width=0.66in,height=0.66in]{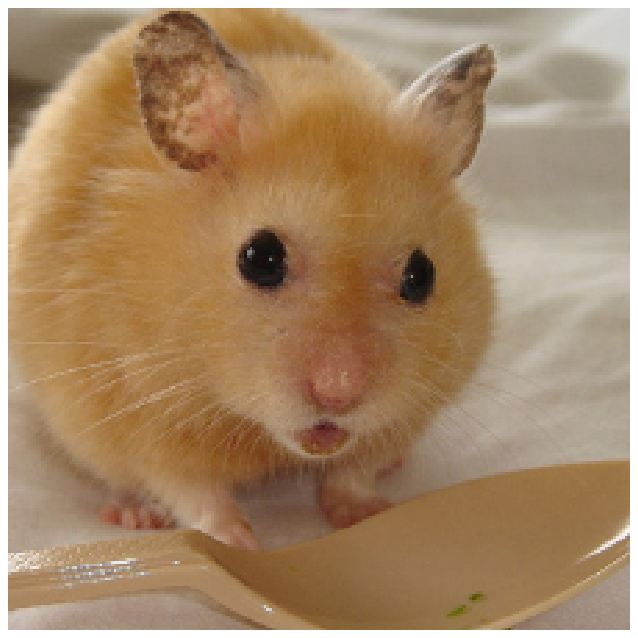}} 
&\textbf{\color{blue}golden hamster, Syrian hamster} & \textbf{\color{blue}golden hamster, Syrian hamster} & \textbf{\color{blue}golden hamster, Syrian hamster}  \\
\rule{0pt}{2.5ex}&rhesus, rhesus monkey & rodent, gnawer &  Eurasian hamster \\
\rule{0pt}{2.5ex}&pipe & Eurasian hamster & prairie dog, prairie marmot \\
\rule{0pt}{2.5ex}&shaker & rhesus, rhesus monkey & skink, scincid, scincid lizard \\
\rule{0pt}{2.5ex}&American mink, Mustela vison & rabbit, coney, cony &  mountain skink\\
\hline
\rule{0pt}{2.5ex}\multirow{5}{*}{\includegraphics[width=0.66in,height=0.66in]{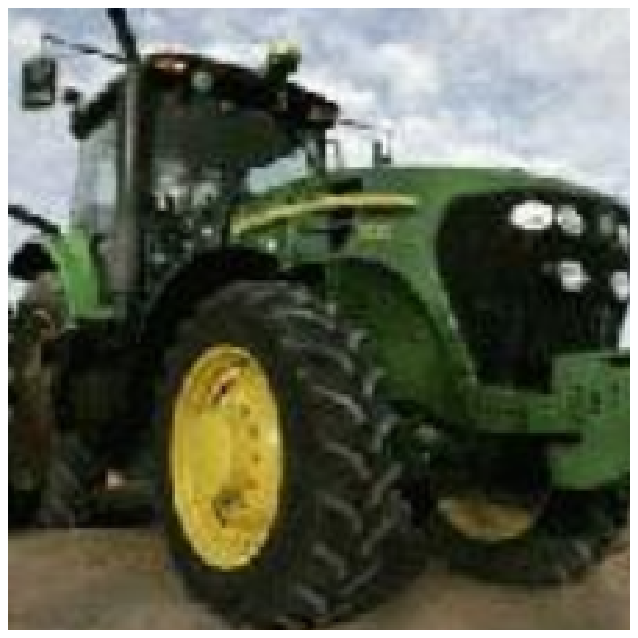}} 
\rule{0pt}{2.5ex}&truck, motortruck & flatcar, flatbed, flat & \textbf{\color{blue}farm machine} \\
\rule{0pt}{2.5ex}&skidder & truck, motortruck & cultivator, tiller \\
\rule{0pt}{2.5ex}&tank car, tank & tracked vehicle & skidder \\ 
\rule{0pt}{2.5ex}&automatic rifle, machine rifle & bulldozer, dozer & bulldozer, dozer  \\
\rule{0pt}{2.5ex}&trailer, house trailer & wheeled vehicle & haymaker, hay conditioner \\
\hline
\rule{0pt}{2.5ex}\multirow{5}{*}{\includegraphics[width=0.66in,height=0.66in]{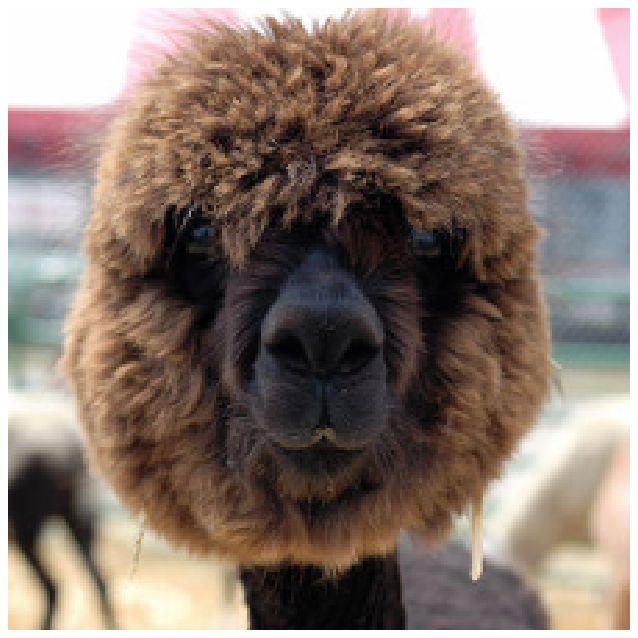}} 
\rule{0pt}{2.5ex}&kernel & dog, domestic dog & mastiff \\
\rule{0pt}{2.5ex}&littoral, litoral, littoral zone, sands & domestic cat, house cat & \textbf{\color{blue}alpaca, Lama pacos} \\
\rule{0pt}{2.5ex}&carillon & schnauzer & domestic llama, Lama peruana  \\
\rule{0pt}{2.5ex}&Cabernet, Cabernet Sauvignon & Belgian sheepdog & guanaco, Lama guanicoe \\
\rule{0pt}{2.5ex}&poodle, poodle dog & domestic llama, Lama peruana & Seeing Eye dog \\
\hline
\end{tabular}\\
\end{center}

\centering
\caption{Top-$k$ accuracy in ImageNet 2011fall zero-shot learning task (\%)}
\begin{tabular}{|c|c|c|l|c|c|c|c|c|}
\hline
\rule{0pt}{2.5ex}Test Set&\#Classes &\#Images &Method & Top-1 &Top-2 &Top-5 &Top-10 &Top-20 \\
\hline 
\rule{0pt}{2.5ex} &&&DeViSE (500-dim)   &0.8 &1.4 &2.5 &3.9 & 6.0\\
\rule{0pt}{2.5ex} &&&ConSE (500-dim)  &1.4 &2.2 &3.9 &5.8  & 8.3\\
\rule{0pt}{2.5ex} &&&Our method (100-dim, random)  & 1.4 & 2.2 & 3.4 & 4.3 & 5.2 \\
\rule{0pt}{2.5ex} &&&Our method (100-dim, PCA)  & 1.6 & 2.7 & 4.6 & 6.4 & 8.6 \\
\rule{0pt}{2.5ex} &&&Our method (100-dim, ICA)  & 1.6 & 2.7 & 4.6 & 6.3 & 8.5 \\
\rule{0pt}{2.5ex} Unseen&20842 &12.9 million &Our method (500-dim, random)  & \textbf{1.8} & 2.9 & 5.0 & 6.9 & 8.8\\
\rule{0pt}{2.5ex} &&&Our method (500-dim, PCA)  & \textbf{1.8} & \textbf{3.0} & \textbf{5.2} & \textbf{7.3} & 9.6\\
\rule{0pt}{2.5ex} &&&Our method (500-dim, ICA)  & \textbf{1.8} & \textbf{3.0} & \textbf{5.2} & \textbf{7.3} & \textbf{9.7}\\
\rule{0pt}{2.5ex} &&&Our method (900-dim, random)  & \textbf{1.8} & \textbf{3.0} & 5.1 & 7.2 & 9.6 \\
\rule{0pt}{2.5ex} &&&Our method (900-dim, PCA)  & \textbf{1.8} & \textbf{3.0} & \textbf{5.2} & \textbf{7.3} & \textbf{9.7}\\
\rule{0pt}{2.5ex} &&&Our method (900-dim, ICA)  & \textbf{1.8} & \textbf{3.0} & \textbf{5.2} & \textbf{7.3} & \textbf{9.7}\\
\hline 
\rule{0pt}{2.5ex} &&&DeViSE (500-dim) &0.3 &0.8 &1.9 &3.2 &5.3 \\
\rule{0pt}{2.5ex} &&&ConSE (500-dim) &0.2 &1.2 &3.0 &5.0 &7.5\\
\rule{0pt}{2.5ex} &&&Our method (100-dim, random)  & \textbf{6.7} & 8.2 & 10.0 & 11.1 & 12.1 \\
\rule{0pt}{2.5ex} &&&Our method (100-dim, PCA)  & \textbf{6.7} & 8.1 & 10.3 & 12.4 & 14.8 \\
\rule{0pt}{2.5ex} &&&Our method (100-dim, ICA)  & \textbf{6.7} & 8.1 & 10.4 & 12.4 & 14.7 \\
\rule{0pt}{2.5ex} Both & 21841 &14.2 million &Our method (500-dim, random)  & \textbf{6.7} & \textbf{8.5} & 11.2 & 13.4 & 15.6 \\
\rule{0pt}{2.5ex} &&&Our method (500-dim, PCA)  & \textbf{6.7} & \textbf{8.5} & \textbf{11.4} & \textbf{13.7} & \textbf{16.3} \\
\rule{0pt}{2.5ex} &&&Our method (500-dim, ICA)  & \textbf{6.7} & \textbf{8.5} & \textbf{11.4} & \textbf{13.7} & \textbf{16.3} \\
\rule{0pt}{2.5ex} &&&Our method (900-dim, random)  & \textbf{6.7} & \textbf{8.5} & \textbf{11.4} & \textbf{13.7} & 16.2 \\
\rule{0pt}{2.5ex} &&&Our method (900-dim, PCA)  & \textbf{6.7} & \textbf{8.5} & \textbf{11.4} & \textbf{13.7} & \textbf{16.3} \\
\rule{0pt}{2.5ex} &&&Our method (900-dim, ICA)  & \textbf{6.7} & \textbf{8.5} & \textbf{11.4} & \textbf{13.7} & \textbf{16.3} \\
\hline                                                 
\end{tabular}
\label{zero_shot_unseen}
\end{minipage}
\end{table*}

\begin{table*}[t!]
\centering
\scriptsize
\caption{Top-$k$ accuracy in ImageNet ILSVRC2012 validation set (\%)}
\begin{tabular}{|c|c|c|l|c|c|c|c|c|}
\hline
\rule{0pt}{2.5ex}Test Set &\#Classes &\#Images &Method & Top-1 &Top-2 &Top-5 &Top-10 &Top-20 \\
\hline 
\rule{0pt}{2.5ex}&&&Softmax baseline (1000-dim) & 55.6& 67.4& 78.5& 85.0 & - \\
\rule{0pt}{2.5ex}&&&DeViSE (500-dim) &53.2 &65.2 &76.7 &83.3 & - \\
\rule{0pt}{2.5ex}&&&ConSE (500-dim) &54.3 &61.9 &68.0 &71.6 & - \\
\rule{0pt}{2.5ex}&&&Our softmax baseline (1000-dim) & \textbf{67.1} &\textbf{78.8} &\textbf{87.9} &\textbf{92.2} & \textbf{95.2} \\
\rule{0pt}{2.5ex}&&&Our method (100-dim, random) & 67.0 & 74.6 &   77.8 & 79.1 & 80.4 \\ 
\rule{0pt}{2.5ex}&&&Our method (100-dim, PCA) & 67.0 & 76.9 &  84.6 & 88.5 & 91.5 \\
\rule{0pt}{2.5ex}Seen &1000 &50000 &Our method (100-dim, ICA) & 67.0 & 76.9 & 84.6 & 88.5 & 91.5 \\
\rule{0pt}{2.5ex}&&&Our method (500-dim, random) & \textbf{67.1} & 77.3 & 83.5 & 85.4 & 86.6 \\ 
\rule{0pt}{2.5ex}&&&Our method (500-dim, PCA) & \textbf{67.1} & 78.2 & 86.2 & 89.4 & 91.2 \\
\rule{0pt}{2.5ex}&&&Our method (500-dim, ICA) & \textbf{67.1} & 78.2 & 86.2 & 89.3 & 91.2 \\
\rule{0pt}{2.5ex}&&&Our method (900-dim, random) & \textbf{67.1} & 78.3 & 86.0 & 88.6 & 90.1\\ 
\rule{0pt}{2.5ex}&&&Our method (900-dim, PCA) & \textbf{67.1} & 78.5 & 86.6 & 89.8 & 91.7 \\
\rule{0pt}{2.5ex}&&&Our method (900-dim, ICA) & \textbf{67.1} & 78.4 & 86.5 & 89.8 & 91.7 \\
\hline                                                 
\end{tabular}
\label{zero_shot_seen}
\end{table*}

In Table \ref{zero_shot_images}, we show the results of the three zero-shot learning methods on the test images selected in \cite{norouzi2013zero}. Same as conSE, our method gives correct or reasonable predictions. 

In Table \ref{zero_shot_unseen}, we show the results of different methods on ImageNet 2011fall. Our method performs better when using PCA or ICA for the visual features than random features. And our method with random, PCA, or ICA features, achieves the state-of-the-art records on this zero-shot learning task.

In Table \ref{zero_shot_seen}. we show the results of different methods on ImageNet ILSVRC2012 validation set of 1000 seen classes. While the goal here is not to classify images of seen classes, it is desirable to measure how much accuracy a zero-shot learning method would lose compared to the softmax baseline. Again, we can see that our method performs better using PCA or ICA for the visual features than random features.

The results show that in our method the PCA or ICA matrix as visual features of object classes performs better than a random matrix. Therefore, these visual features, learned PCA and ICA on the outputs of CNN, are indeed effective for the subsequent tasks. The results also show that PCA and ICA give the essentially same classification accuracy. Therefore, in practice we can use PCA instead of ICA, which has much higher computational costs. For a more comprehensive discussion on PCA vs. ICA for recognition tasks, see \cite{asuncion2007equivalence}.
The code for reproducing the experiments is in \\

\url{https://github.com/yaolubrain/ULNNO}

\newpage

\section{Discussion and Conclusion}
The outputs of a neural network contains rich information. 
It has been claimed that one can determine a neural network architecture by observing its outputs given arbitrary inputs \cite{fefferman1994recovering}.
Also, it has been shown that one can reconstruct the whole image to some degree with only its CNN outputs \cite{dosovitskiy2015inverting}. And smooth regularization on the output distribution of a neural network can help in reducing generalization error in both supervisd and semi-supervised settings \cite{miyato2015distributional}.

CNN achieves the state-of-the-art results on many computer vision tasks such as image classification and object detection. However, despite many efforts of visualizing and understanding CNN \cite{zeiler2014visualizing,simonyan14deep,zhou2014object}, it still reminds a black-box method.  In this paper, we attempted to understand CNN by unsupervised learning. CNN was trained with only one-hot targets, which means we assumed object classes are equally similar. We never told CNN which classes more similar. But unsupervised learning on CNN outputs reveals the visual similarity of object classes. We hope this finding can shed some lights on the object representation in CNN.

We also showed that there is a gap between the visual similarity of object classes in CNN and the semantic similarity of object classes in our knowledge graph. Therefore, a bridge should be built, in order to achieve consistent mapping between visual and semantic representations.

Supervised learning alone cannot deal with unseen classes since there is no training data. By using external knowledge and unsupervised learning algorithms, we can leverage supervised learning so as to make reasonable predictions on the unseen classes while maintaining the compatibility with the seen classes, that is, zero-shot learning. In this paper, we proposed a new zero-shot learning method, which achieves the state-of-the-art results on the ImageNet of over 20000 classes.

\section*{Acknowledgements}
Yao Lu thanks his supervisors Aapo Hyvarinen and Michael Gutmann for their kind guidance. He also thanks Yoshua Bengio for an inspiring conversation on unsupervised learning and Kyunghyun Cho for helpful comments on the paper.

\bibliographystyle{named}
\bibliography{cnn_ica}

\end{document}